# Automated Postediting of Documents

Kevin Knight and Ishwar Chander
USC/Information Sciences Institute
4676 Admiralty Way
Marina del Rey, CA 90292
{knight,chander}@isi.edu

## Abstract

Large amounts of low- to medium-quality English texts are now being produced by machine translation (MT) systems, optical character readers (OCR), and non-native speakers of English. Most of this text must be postedited by hand before it sees the light of day. Improving text quality is tedious work, but its automation has not received much research attention.

Anyone who has postedited a technical report or thesis written by a non-native speaker of English knows the potential of an automated postediting system. For the case of MT-generated text, we argue for the construction of postediting modules that are portable across MT systems, as an alternative to hardcoding improvements inside any one system. As an example, we have built a complete self-contained postediting module for the task of article selection (*a*, *an*, *the*) for English noun phrases. This is a notoriously difficult problem for Japanese-English MT. Our system contains over 200,000 rules derived automatically from online text resources. We report on learning algorithms, accuracy, and comparisons with human performance.

## Automated Postediting

Fully automatic, high-quality translation is still an elusive goal for broad-coverage natural language processors. Current machine translation (MT) systems usually employ a human posteditor to transform the MT output into usable, quality text. If the posteditor can do this transformation in less time than it takes to translate from scratch, then the MT system is economically viable. Many commercial systems exist on this principle.

Improving a particular MT system often means automating something that the posteditor is doing. The system gets further, leaving the posteditor with less to do. And usually, the improvements are coded into the internals of the MT system, becoming part of a black box.

Another way to think about automating postediting tasks is to build automated postediting modules that are detachable and independent of any particular MT system. Figure 1 shows this distinction. The advantage of detachable posteditors is that they are portable across MT systems. They accomplish their tasks without reference to the internal algorithms and representations of particular systems. With portability comes leverage: one piece of coded linguistic analysis can be used to improve the quality of many automatic translators. Furthermore, postediting modules can clean up text generated by humans (whose internal algorithms are unknown). Texts that are imperfectly scanned by optical character readers (OCR) are also grist for automated postediting.

We envision two types of postediting modules. One type is *adaptive*, the other *general*. The rest of this section briefly discusses what the former type would look like; the remainder of the paper describes a posteditor of the latter type that has been designed, built, and tested.

The idea of an adaptive posteditor is that an automatic program can watch a human postedit documents, see which errors crop up over and over (these will be different for any given system/domain pair), and begin to emulate what the human is doing. As yet, no adaptive posteditors, portable across systems and domains, have been built for MT. One place to start would be a large corpus of "pre-postedited" text aligned with corresponding postedited text. Statistical machine translation techniques could then be applied to learn the mapping. One would hope that such techniques would have an easier time learning to translate bad English to good English than, say, Japanese to good English.

## Article Selection

A general posteditor must be useful for improving text produced by a wide variety of MT systems and non-native speakers and should operate equally across all domains. While working on Japanese-English translation within the PANGLOSS project (Nirenburg & Frederking 1994; Knight & Luk 1994), we have constructed an automatic posteditor for inserting articles (*a*, *an*, and *the*) into English text. Several factors mo-

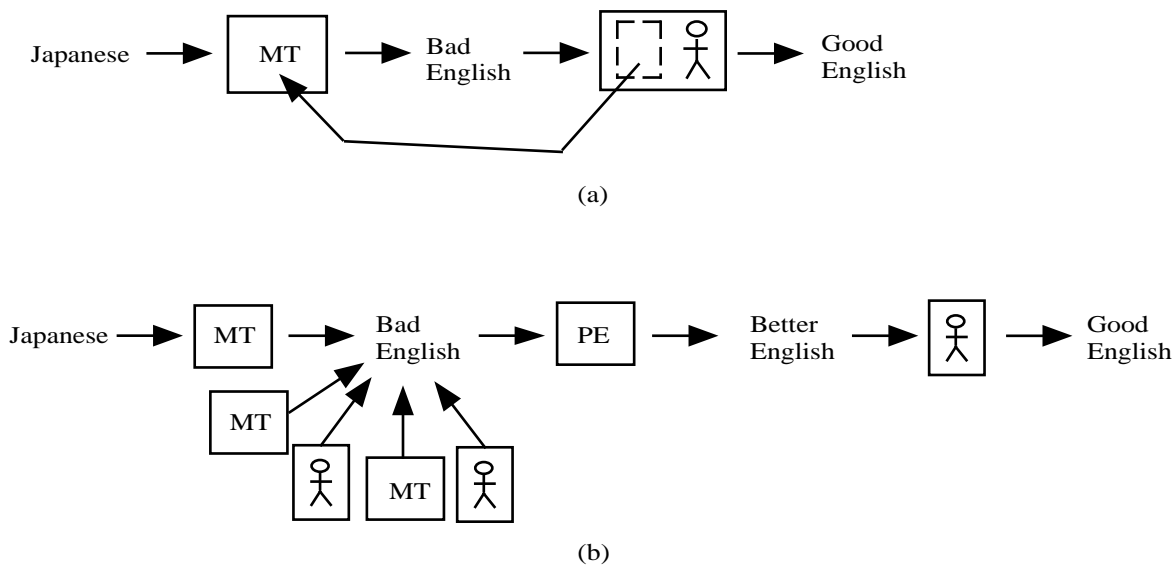

Figure 1: Two views of automating tasks of postediting. In (a), posteditor work is automated and moved into a particular MT system. In (b), a detached postediting module is created. It serves to improve the quality of text produced by several MT systems and non-native speakers.

tivated this choice:

- The Japanese language has no articles, but article-free English is difficult to read.
- Inserting articles is tedious work for a human English-speaking posteditor.
- Non-native English speakers find accurate article selection very difficult, even after years of practice.
- Doing the task well is beyond the capabilities of current automatic grammar checkers.

Here is an example of the task. The following text comes in without articles:

```
Stelco Inc. said it plans to shut down three
Toronto-area plants, moving their fastener
operations to leased facility in Brantford,
Ontario.
Company said fastener business "has been
under severe cost pressures for some time."
Fasteners, nuts and bolts are sold to North
American auto market.
Company spokesman declined to estimate impact
of closures on earnings. He said new facility
will employ 500 of existing 600 employees.
Steelmaker employs about 16,000 people.
```

The posteditor transforms this text as follows:

```
Stelco Inc. said it plans to shut down three
Toronto-area plants, moving their fastener
operations to a leased facility in Brantford,
Ontario.
The company said the fastener business "has
been under severe cost pressures for some time."
The fasteners, nuts and bolts are sold to the
North American auto market.
A company spokesman declined to estimate the
impact of the closures on earnings. He said
the new facility will employ 500 of the
existing 600 employees. The steelmaker employs
about 16,000 people.
```

Accuracy on this task can be measured quantitatively. As a first approximation, we take newspaper-quality English text, remove the articles, have our program re-insert articles, and compare the resulting text to the original. When the two match (i.e., the program has restored the original article), we score a success. Otherwise, we score a failure. Dividing successes over total articles in the text yields accuracy. Note that this scoring method is a bit unforgiving: if either article is permissible in a certain phrase, we still score a failure for not matching the article chosen by the author of the original text. We pay this price in order to get a fully automatic evaluation set-up. Real accuracy figures will be slightly higher than those reported.

We have made two simplifications to the problem of article selection for the purposes of the experiments described in this paper. One is that we assume noun phrases are already marked as singular or plural, as in normal English. Japanese has no such markings, however, so it would be more realistic to build a posteditor to insert plural forms and articles simultaneously. The system described here is more suited to Russian-English translation, since Russian has plurals but no articles. The second assumption we make is that we are given placeholders for the articles, and each decision is a binary one: *the* versus *a/an*.[1] We are now in the process of lifting these restrictions in order to select plurals and other types of articles, especially the *zero* article.

## Performance Bounds

What expectations should we have about how well a program might perform article selection? We performed several experiments to answer this question.

Guessing by coin-flip (heads equals *the*, tails equals *a/an*) yields an accuracy of 50%. But we can improve our accuracy to about 67% simply by guessing *the* every time. We determined this by inspecting 40 megabytes of Wall Street Journal text, noting the breakdown of articles as follows:

- $a = 28.2\%$
- $an = 4.6\%$
- $the = 67.2\%$

So 67% is a good lower bound on expected performance; we shouldn't do worse.

To get some upper bounds, we tested two human subjects on the following task: Given an English text with articles replaced by blanks, try to restore the original articles.[2] Subjects performed with accuracies between 94% and 96%. These numbers show that articles contain very little information (in the Shannon sense), because they are quite predictable from context. This confirms our intuition that languages without articles transmit information at no special handicap. But English articles are not completely predictable, and 95% is a good upper bound on performance. An analysis of the 5% errors shows that some are cases where *a* and *the* are synonymous in context, while others are cases where the human subject failed to read the author's intent.

The human subjects were asked to perform two other tasks. In one, subjects had to predict articles in a very limited context, namely, given just the head noun following the article and its premodifiers. Here is a sample of this task:

```
(??? "reduced" "dividend")
(??? "fiscal" "year")
(??? "profit")
(??? "new" "dividend" "rate")
(??? "pilot" "training" "school")
```

```
(??? "price")
```

Performance was between 79% and 80%, thirteen percentage points better than no-context performance (always guess *the*) but fifteen percentage points worse than full-context performance. This gives us some idea of local versus discourse effects in article selection.[3]

The final task used a slightly expanded context. Subjects were shown the core noun phrase plus two words to the left of the unknown article and two words to the right of the head noun, e.g.:

```
(("It" "said")
 (??? "reduced" "dividend")
 ("reflects" "the"))

(("losses" "for")
 (??? "fiscal" "year")
 ("ending" "Oct"))

(("last" "had")
 (??? "profit")
 ("in" "1985"))

(("1985" ".")
 (??? "new" "dividend" "rate")
 ("is" "payable"))

(("Academy" ",")
 (??? "pilot" "training" "school")
 ("based" "at"))

(("to" "disclose")
 (??? "price")
 ("." "Comair"))
```

Subjects achieved an accuracy of 83% to 88%. This amount of context is what we showed our program when we evaluated it, so these figures are good upper bounds for its behavior.

To summarize the results of this section:

|   | human | machine |
|---|---|---|
| random | 50% | 50% |
| always guess *the* | 67% | 67% |
| given core NP context | 79-80% |   |
| given NP plus 4 words | 83-88% | ? |
| given full context | 94-96% |   |

These numbers are not accurate to any degree of statistical significance, due to the small survey size. It was not our intention to make a full-blown psychological study of human article selection; rather, we wanted to establish rough targets to tell us how far we have to go and when to stop.

---

[1] Choosing between *a* and *an* is done as a final step. We use a hand-built trie-driven algorithm detailed enough to distinguish *a unique guest* from *an unimpressed guest*. Minor difficulties remain only with previously unseen acronyms, e.g., distinguishing *a NATO grant* from *an NIH grant*.

[2] This blank is an ambiguous "pseudo-word" in the sense of (Gale, Church, & Yarowsky 1992), which we proceed to disambiguate.

[3] Noun phrases were presented to human subjects in a random order. Presenting them in original-text order would allow the humans to make use of the same discourse effects we are trying to screen out.

## Algorithms

In thinking about the selection of articles *a*, *an*, and *the*, natural starting points are the notions of definiteness and discourse. Articles *a* and *an* introduce new discourse entities, while *the* often signals back to previously mentioned or inferred entities. As the previous section indicated, some of article selection requires full discourse context, while some requires only local context. Usage rules are extensively covered in 100 pages of the English grammar reference (Quirk & Greenbaum 1973). These include:

- General knowledge: *the* President, *the* Moon.
- Immediate situation: feed *the* cat.
- Indirect anaphora: he bought a car, but *the* engine was faulty.
- Sporadic reference: she goes to *the* theater every week.
- Logical uniqueness: they have *the* same hobby.
- Body parts: hit in *the* eye.
- Generic use: *the/a* tiger is a ferocious beast.
- Referential, Non-uniqueness: *a* dog bit me.
- Nonreferential, Description: she is *a* good player.

These rules (in spelled-out form) are proper analyses of article selection, but they are difficult to operationalize, due to the representations and world knowledge required. Coding large amounts of general knowledge with AI techniques is known to be hard. And even fully armed with such knowledge, non-native speakers still have trouble mastering the rules.[4]

The rules are generally easier to operationalize when they are accompanied by examples and exceptions, the more the better. Exceptions are common, as demonstrated by differences between British and American article usage. This led us to leave the rules behind and move to a purely data-driven approach.

We generated a database of over 400,000 core noun phrases from Wall Street Journal text. This database is the same one we used to test the human subjects:

---

[4] Notice also that most rules describe when to use *the*. For *a* and *an*, (Quirk & Greenbaum 1973) say: "The indefinite article is notionally the 'unmarked' article in the sense that it is used (for singular count nouns) where the conditions for the use of *the* do not obtain."

Unfortunately, with *the* occurring 67% of the time, we are in much more need of finding conditions for the use of *a* and *an*. To see this, consider a *the* rule that correctly covers 25% of all noun phrases. The remaining 75% would break down as follows: 56% *the* and 44% *a/an*. In the absence of other rules, we might as well also guess *the* for these noun phrases, yielding the same overall accuracy of 67%. On the other hand, an *a/an* rule covering 25% of all noun phrases would leave 89% *the* and 11% *a/an*. Guessing *the* for these would yield a total accuracy of 92%.

```
(("It" "said")
 ("the" "reduced" "dividend")
 ("reflects" "the"))

(("losses" "for")
 ("the" "fiscal" "year")
 ("ending" "Oct"))

(("last" "had")
 ("a" "profit")
 ("in" "1985"))

(("1985" ".")
 ("The" "new" "dividend" "rate")
 ("is" "payable"))

(("Academy" ",")
 ("a" "pilot" "training" "school")
 ("based" "at"))

(("to" "disclose")
 ("the" "price")
 ("." "Comair"))
```

We then looked at which words, word features, and combinations of these were most predictive of the given article. The head noun is most critical. If the head noun is *White House* (this happened 238 times in the database), the article is almost always *the* (236/238 times). Plural head nouns, regardless of root, usually have *the*. Premodifiers like *next* and *same* prefer *the*, as do superlative adjective premodifiers, regardless of root. Combinations are critical: *deficit* as a head noun is not very predictive, and neither is *Federal* as a premodifier, but combined, they have a strong preference. If *ago* follows the head noun directly (this occurred 881 times), the article is never *the*. Words like *triple* are influential just before the article, as in *triple the cost*. Other frequent patterns include *rest of the*, *clear the way*, *X% a year*, *a sign of*, etc.

This exploratory data analysis led us to develop a set of binary features that characterize any noun phrase. These features are either lexical (*word before article is 'triple'*) or abstract (*word after head noun is a past tense verb*). Lexical features were obtained directly from the database. Any word appearing in a given position more than once counts as a feature. Abstract features include part-of-speech, plural marking, tense, and subcategory (superlative adjective, mass noun, etc.).

The problem now is to predict the article based on the features of the context. Given that the features are not independent, care must be taken to integrate the "votes" each feature wants to make. We decided to take a decision tree approach (Quinlan 1986; Breiman et al. 1984) to modeling feature interaction.

To the decision tree builder, each feature $f$ has three statistics of interest. The first is frequency of occurrence $p_1$, the second is the distribution of *the* versus

*a/an* for noun phrases in which the feature is present $p_2$, and the third is the distribution for those without the feature $p_3$. Choosing a feature splits the data. The goal is a roughly even split with the resulting two data sets being more informative than the original. An information-theoretic approach to choosing the best feature is to pick feature $f$ that minimizes:

$$-p_1 \cdot \log(p_2) - (1 - p_1) \cdot \log(p_3)$$

The tree builder recurses on the data sets that result from the feature-based split. We terminate the recursion when the training examples are 98% in agreement with one another. Applying the tree to a new noun phrase means walking down the tree, based on the phrase's features, returning the article stored at the leaf node.

The main difficulty with learning is that we have over 400,000 training examples and over 30,000 features. (The number of features is high because it includes lexical features such as *word directly after head noun is 'ago'*.) Choosing a feature for the root of the decision tree would require, by the straightforward implementation, 400,000 · 30,000 operations to compute $p_1$, $p_2$, and $p_3$ for each feature. To cut down on the computation, we throw out features with less than four instances. We also note that any given lexical feature will have a small $p_1$, i.e., it will occur infrequently in the database. So we can compute a feature's $p_3$ by closed form, given a distribution $q$ at the node:

$$p_3 = \frac{q - p_1 \cdot p_2}{1 - p_1}$$

We then index features with associated training instances for fast computation of $p_1$ and $p_2$.

## Results

Because the space of features and training examples is still large, we decided to break the training data into subsets. We began with high-frequency head nouns. For example, *president* occurs 1420 times in our database. The breakdown of associated articles is as follows:

- *a/an* = 46.5%
- *the* = 53.5%

We trained a decision tree to generate articles for *president* noun phrases based on the premodifiers, two words before the article, and two words after. Training on 90% of the data and testing on 10% yielded a tree of 171 questions and a test set accuracy of 89%. Figure 2 shows the learning curve for noun phrases ending with *president*. We built ten decision trees, giving each tree more training data than the last. The curve shows test-set accuracy for each tree, along with tree size.

Performance on noun phrases ending in *year* was 94%, and for *stock* 90%.

These good scores indicate that with enough training data, we can generate highly accurate decision trees.

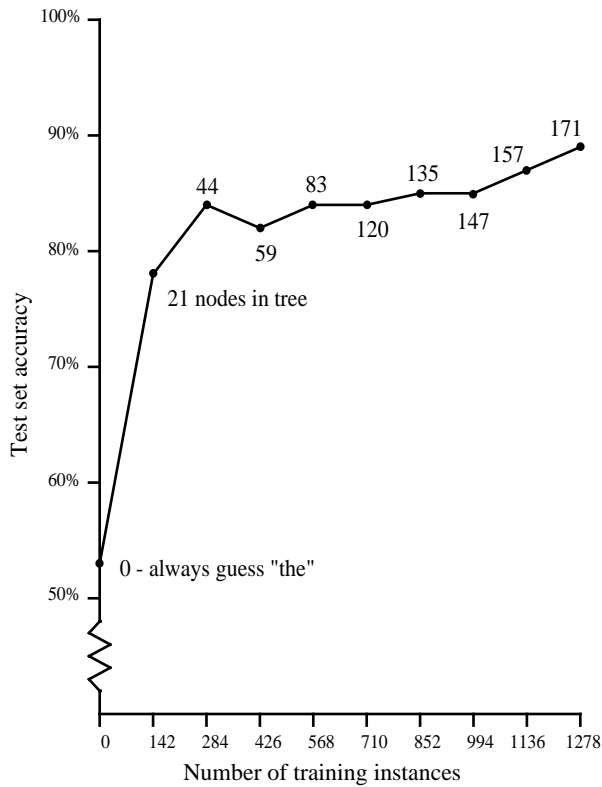

Figure 2: A learning curve for article selection. The graph shown is for noun phrases ending in *president*. The curve shows test-set performance plotted against training instances fed into the decision tree builder. Data points are annotated with the size of the decision tree built.

Unfortunately, of the 23,871 distinct head nouns found in our training database, most occur only once or twice. But—fortunately—the 3413 head nouns occurring at least 25 times account for 84% of the instances.

At the time of this writing, we have built 1600 trees for the 1600 most popular head nouns, covering 77% of the test-set instances. On these instances, we achieve 81% accuracy, which approaches human performance. For the remaining 23%, we simply guess *the*, for an accuracy of 66%. The overall accuracy is 78%. We expect to improve this figure by several points by building more trees, adding more instances, and aggregating many of the low-frequency head nouns on the basis of shared features.

## Related Work and Discussion

Definiteness/indefiniteness of noun phrases has been an object of linguistic study for a long time. Computational linguists have been particularly concerned with finding referents for definite noun phrases (Grosz, Joshi, & Weinstein 1983; Sidner 1983), an anaphora problem with clear applications for text understanding.

Text generation research has tackled problems in generation of determiners from a semantic representation (see, e.g., (Elhadad 1993) for a discussion of judgment determiners). Practical MT systems deal with article selection when the source language does not have articles, but the target language does.

Our postediting algorithm achieves an accuracy rate of 78% on financial texts without the benefit of a semantic representation to work from. It does not require an analysis of the source language text, nor even that such text exists. If such representations were available, we could trade portability for increased accuracy.

In some sense, our results are another testament to the power of "know-nothing" statistics to achieve reasonable accuracy and broad coverage. But this is an exaggeration, because our program has a great deal of knowledge built into it. The biggest piece is the noun phrase parser, which finds the most predictive element of the context. Often the head noun is far from the article, and often other nouns intervene—for these reasons, features like "noun x appears within 3 words to the right of the article" are too imprecise.

Word classes also form an important piece of knowledge. Abstract features that characterize large numbers of training cases provide us with trustworthy statistics. Lexical features occurring infrequently may be important, but it is impossible to distinguish them from noise. At present, we only use syntactic word classes, but semantic classes (Miller 1990; Brown *et al.* 1992) could help to alleviate our sparse data problem.

## Future Work

As noted earlier, we are now working to relax certain assumptions about the article selection task and our evaluation. We are extending our training database to include noun phrases with *some/any* and the *zero* article. We will re-run our human experiments to get new upper bounds and use the same training procedure (with binary features) but distributions over four possible outcomes: *a/an*, *the*, *some/any*, and *zero*. We do not intend to produce determiners *this* and *that*, as these are explicitly marked in Japanese. The new database is being built from part-of-speech-tagged text; our current algorithms parse noun phrases anchored at *a*, *an*, and *the* without the need for tagging software. We also plan to:

- Measure how performance degrades on other types of text. Currently, we test the algorithm on text drawn from the same population it was trained on.

- Incorporate full-context discourse features into the model. Humans do about 15% better when the full discourse is available. With the right features, we hope to capture some of this gain. What the right features are, however, is still unclear.

- Turn to other routine postediting tasks. These include plural selection, preposition selection, and punctuation.

## Acknowledgements
We would like to thank Eduard Hovy for his support and for comments on a draft of this paper. Thanks also to Yolanda Gil for her comments. This work was carried out under ARPA Order No. 8073, contract MDA904-91-C-5224.